\pdfoutput=1
\documentclass[11pt]{article}

\usepackage[preprint]{acl}

\usepackage{times}
\usepackage{latexsym}
\usepackage{adjustbox}
\usepackage{graphicx}
\usepackage{makecell}
\usepackage{algorithm}
\usepackage{algorithmic}
\usepackage{tabularx}
\usepackage{booktabs}
\usepackage{multirow}
\usepackage{makecell}
\usepackage{array} 
\usepackage{xcolor}     % 支持 \textcolor
\usepackage{pifont} % already needed for newcolumntype
\usepackage{ragged2e}    % for \RaggedRight
\usepackage{amssymb}
\newcolumntype{Y}{>{\raggedright\arraybackslash}X}

\usepackage[T1]{fontenc}
\usepackage{CJKutf8} % Added for Chinese language support
\usepackage[utf8]{inputenc}

\usepackage{microtype}

\usepackage{inconsolata}

\usepackage{amsmath}

\usepackage{graphicx}

\title{CCL-XCoT: An Efficient Cross-Lingual Knowledge Transfer Method for Mitigating Hallucination Generation}

\author{%
    \textbf{Weihua Zheng,\quad Roy Ka-Wei Lee, \quad Zhengyuan Liu,\quad}\\
    \textbf{Kui Wu,\quad AiTi Aw,\quad Bowei Zou}\\[2mm]
    Institute for Infocomm Research (I$^2$R), A*STAR, Singapore\\[1mm]
    {\small \texttt{\{zheng\_weihua, liu\_zhengyuan, wuk, aaiti, zou\_bowei\}@i2r.a-star.edu.sg, s.roylee@gmail.com}}
}

\date{}

\begin{document}
\begin{CJK}{UTF8}{gbsn} % Start Chinese language environment
\maketitle
\begin{abstract}
Multilingual Large Language Models (MLLMs) demonstrate strong generalization across languages, yet they remain prone to hallucinations, especially in low-resource languages, due to training data imbalances. These hallucinations, which include inaccurate or fabricated outputs, are particularly problematic in domain-specific generation tasks \cite{chataigner2024multilingualhallucinationgapslarge}. To address this challenge, we propose \textsf{CCL-XCoT} (Curriculum-based Contrastive Learning-based Cross-lingual Chain-of-Thought), a two-stage fine-tuning framework for mitigating hallucination in MLLMs. Our approach first enhances cross-lingual semantic alignment through curriculum-based contrastive learning combined with next-token prediction during continued pre-training. Building on this foundation, we then introduce a cross-lingual Chain-of-Thought (\textsf{XCoT}) prompting strategy during instruction fine-tuning, which guides the model to reason in a high-resource language before generating answers in the target low-resource language. Experimental results show that \textsf{CCL-XCoT} reduces hallucination rates by up to 62\% and substantially improves factual knowledge transfer across language pairs, without relying on external retrieval or multi-model ensembles.
\end{abstract}

\section{Introduction}
Despite the impressive performance of large language models (LLMs) across a wide range of natural language processing tasks, hallucination remains a fundamental and unresolved challenge—particularly in multilingual settings. This issue is especially pronounced in low-resource languages, where limited training data increases the likelihood of inaccurate or misleading outputs. Hallucinations in LLMs typically manifest in three forms: (1) \textit{context-related hallucinations}, which contradict general world knowledge; (2) \textit{self-conflicting hallucinations}, involving internal inconsistencies such as flawed reasoning; and (3) \textit{ungrounded hallucinations}, where the generated content diverges from source facts despite maintaining surface fluency \cite{lei2023chainnaturallanguageinference}. This work focuses on mitigating context-related hallucinations in low-resource languages.

A major contributor to context-related hallucinations is the asymmetry in knowledge coverage between high-resource and low-resource languages. Recent studies have shown that multilingual models often fail to share factual knowledge across languages \cite{hu2025largelanguagemodelscrosslingual,schut2025multilingualllmsthinkenglish}. This limitation is partially attributed to the language-agnostic next-token prediction (NTP) pretraining objective, which does not explicitly enforce cross-lingual semantic alignment. As a result, multilingual LLMs struggle to generalize factual knowledge to underrepresented languages.

Existing solutions such as Retrieval-Augmented Generation (RAG) have demonstrated effectiveness in English, but perform poorly in low-resource settings due to unreliable retrieval and limited language coverage \cite{niu2024ragtruthhallucinationcorpusdeveloping}. Other strategies, such as Chain-of-Thought (CoT) prompting \cite{Ayala_2024,song-etal-2024-rag,sun2025redeepdetectinghallucinationretrievalaugmented,li2024banishingllmhallucinationsrequires}, improve factual accuracy via step-by-step reasoning. However, CoT assumes the model already possesses sufficient internal knowledge in the target language—an assumption that often breaks down in low-resource conditions.

To address these gaps, we propose \textsf{CCL-XCoT}, a two-stage framework for improving factual generation in low-resource languages without relying on external retrieval. Our approach comprises two key components: (i) a \textit{curriculum-based contrastive learning} strategy combined with next-token prediction during continued pretraining to align semantic spaces between high- and low-resource languages, and (ii) a cross-lingual Chain-of-Thought strategy during instruction fine-tuning, which enables models to reason in a high-resource language (e.g., English) before generating answers in the target low-resource language. This design facilitates the transfer of \textbf{non-language-specific} knowledge, such as common facts and historical information, across languages.

This paper makes the following contributions: (i) We propose \textsf{CCL-XCoT}, a two-stage framework that reduces hallucinations in low-resource languages by up to 62\%, without requiring retrieval systems or ensemble models. (ii) We introduce curriculum-based contrastive learning to align multilingual semantic spaces, improving factual understanding and yielding up to 20\% gains on cross-lingual NLU tasks. (iii) We develop \textsf{XCoT} strategy that transfers reasoning patterns from high-resource to low-resource languages, enhancing answer accuracy and completeness. (iv) Our layer-wise analysis reveals that mid-level transformer layers are key to effective cross-lingual knowledge transfer, guiding future efficient finetuning.

\begin{figure*}[t]
  \centering
  \includegraphics[width=0.8\textwidth]{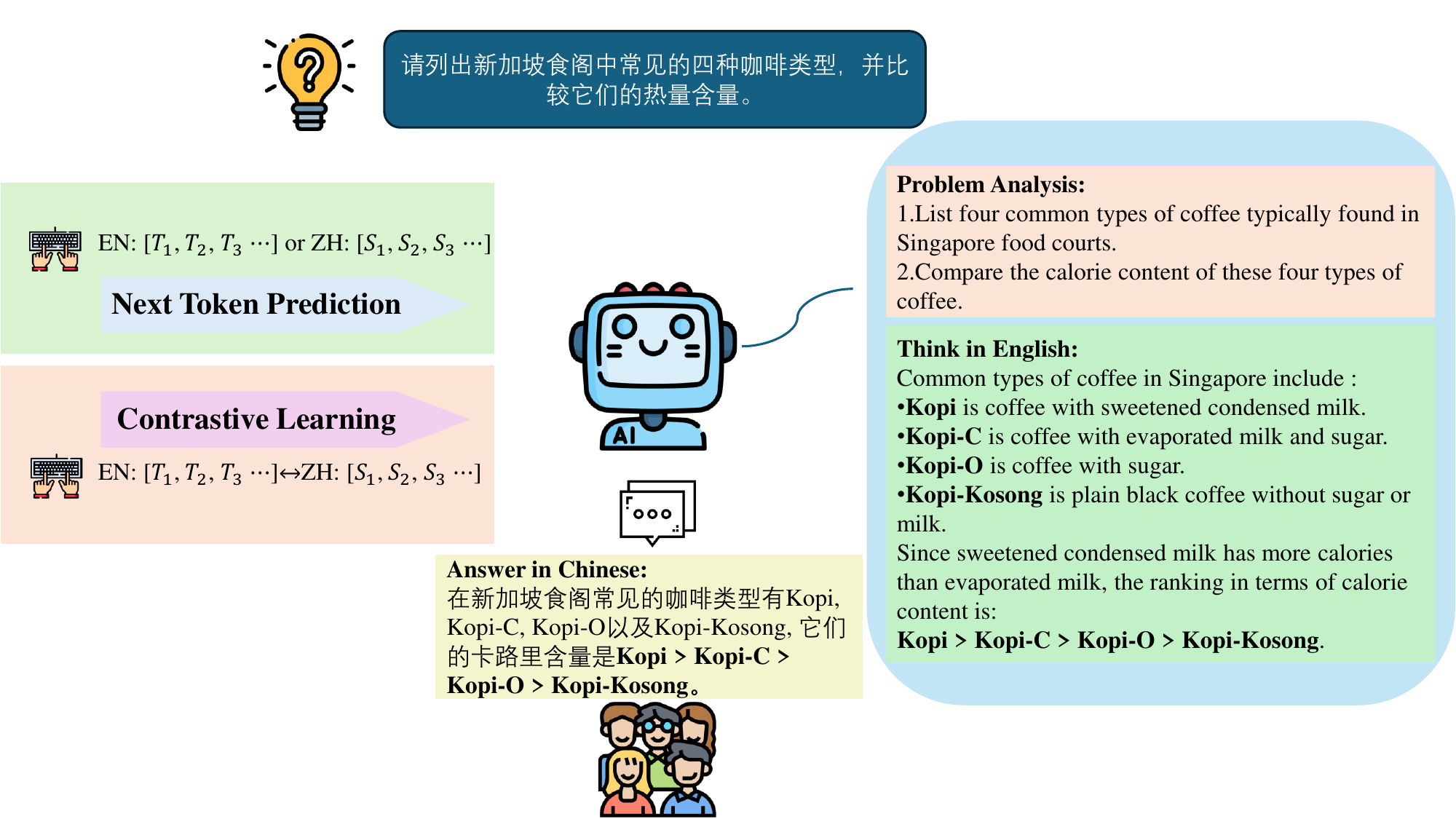}
  \caption{Overview of our proposed framework. The model is pre-trained with a combination of next-token prediction and a two-stage curriculum-based cross-lingual contrastive learning approach to align semantic spaces across languages. During the instruction fine-tuning phase, a cross-lingual CoT strategy guides the model to leverage knowledge in a high-resource language to improve responses in low-resource languages.} 
  \label{fig: Method}
\end{figure*}

\section{Related Work}
LLMs have achieved strong performance in text generation tasks, but remain prone to hallucinations that undermine output reliability. To mitigate this, a range of strategies have been proposed.

\textbf{Retrieval-Based Hallucination Mitigation.} RAG integrates external knowledge to improve LLM outputs' factual accuracy \cite{izacard2022atlasfewshotlearningretrieval, lewis2021retrievalaugmentedgenerationknowledgeintensivenlp, ram-etal-2023-context}. For instance, \citet{Ayala_2024} retrieved relevant task templates using MiniLM embeddings, while \citet{song-etal-2024-rag} proposed Hallucination Aware Tuning(HAT), which uses hallucination detection and GPT-4-based revision to build a preference dataset for Direct Preference Optimization(DPO). Despite promising results in English, RAG remains less effective in low-resource languages due to poor retrieval and limited training coverage \cite{niu2024ragtruthhallucinationcorpusdeveloping}. Moreover, RAG-based systems may generate conflicting outputs if the retrieved context is misinterpreted.

\textbf{Reasoning-Based Mitigation and Self-Correction.} Another line of work explores enhancing internal reasoning to suppress hallucinations without external retrieval. \citet{lei2023chainnaturallanguageinference} introduced CoNLI, which applies Natural Language Inference (NLI) to detect and rewrite unsupported generations. Chain-of-Verification \cite{dhuliawala2023chainofverificationreduceshallucinationlarge} prompts the model to verify its own claims via question-answering before producing final outputs. \citet{ji-etal-2023-towards} adopted a self-reflective, multi-turn generation process for iterative refinement. While effective, these methods rely heavily on robust reasoning and world knowledge, capabilities that often degrade in low-resource languages, making such approaches less generalizable in multilingual contexts.

\textbf{Ensemble and Agreement-Based Methods.}
\citet{wei2024measuringreducingllmhallucination} propose FEWL, a multi-model ensemble that assigns weights to reference LLMs based on their reliability, and introduces a laziness penalty to penalize superficial responses. While FEWL shows promise, it depends on the availability of strong evaluation models, which are often lacking for non-English languages—thereby limiting its cross-lingual effectiveness.

\textbf{Translation-based Methods.} \citet{chai2401xcot} and \citet{lin2024crossin} used machine translation to align semantic spaces by generating code-switch sentences or swapping full question-answer sequences. However, these methods risk cumulative translation errors and ambiguities from fragmented replacements.

\section{Methodology}
We propose a framework to improve decoder-only language models for low-resource languages by integrating NTP, two-stage cross-lingual contrastive learning, and CoT reasoning. Unlike standard causal language modeling, our method explicitly aligns multilingual semantic spaces through contrastive learning and transfers knowledge from high-resource to low-resource languages via CoT prompting. An overview of the framework is illustrated in Figure~\ref{fig: Method}.

\subsection{Background}

\paragraph{Next-Token Prediction in Causal Language Modeling.} 
Causal language models (CLMs) are typically trained using the NTP objective, where each token is predicted based on its preceding context \cite{radford2018improving, radford2019language}. However, this objective is applied primarily on monolingual corpora without enforcing cross-lingual semantic alignment. As a result, multilingual CLMs often exhibit fragmented representations and struggle to generalize knowledge across languages.

To address this, we incorporate contrastive learning during continued pretraining to encourage cross-lingual semantic alignment and improve multilingual transfer. At the same time, continued training may cause catastrophic forgetting of previously acquired knowledge \cite{mccloskey1989catastrophic, ratcliff1990connectionist, gururangan2020don, ke2022continual, liu2019roberta}. To mitigate this, we retain the NTP objective during pretraining and jointly optimize it alongside contrastive learning.

\paragraph{Contrastive Learning for Cross-Lingual Alignment.} 
Contrastive learning has shown promise for aligning semantic spaces, particularly in encoder-decoder models \cite{lewis2019bart, raffel2023exploringlimitstransferlearning, liu2021simcls}. However, its application to decoder-only models remains limited and is largely restricted to English \cite{jain2022contraclm, su2022contrastive}. Its potential for cross-lingual alignment in such models is still underexplored.

To address this, we compute contrastive loss at the sequence level—rather than token level—to accommodate varying tokenization granularities across languages. For two semantically equivalent sentences $S_{m}^{i}$ and $S_{n}^{i}$ from languages $m$ and $n$, we treat them as a positive pair. Batches for each language are defined as:

\begin{align}
B_{m} &= \left[ S_{m}^{i}, S_{m}^{i+1}, \dots, S_{m}^{i+T} \right], \\
B_{n} &= \left[ S_{n}^{i}, S_{n}^{i+1}, \dots, S_{n}^{i+T} \right]
\end{align}
where sentences with matching indices are positives, and all others are treated as negatives.

\subsection{Two-Stage Curriculum-Based Cross-Lingual Contrastive Learning}
To align semantic representations across languages in decoder-only CLMs, we propose a two-stage curriculum-based contrastive learning framework, which progressively enhances cross-lingual understanding by addressing the scarcity of paragraph-level bilingual data and accommodating varying tokenization granularities across languages.

\paragraph{Stage 1: Sentence-Level Contrastive Pretraining.} 
The first stage leverages large-scale sentence-aligned parallel corpora. Given bilingual sentence pairs $\{(S_m^i, S_n^i)\}_{i=1}^T$ from languages $m$ and $n$, we compute sentence embeddings using the final hidden state of the decoder. Each aligned pair $(S_m^i, S_n^i)$ is treated as a positive example; all others in the batch are treated as negatives. The contrastive loss is:

\begin{equation}
\begin{aligned}
\mathcal{L}_{\text{contrastive}} = \frac{1}{T} \sum_{i=1}^T -\bigg( \log \frac{\exp(\text{sim}(S_m^i, S_n^i) / \tau)}{\sum\limits_{j=1}^{T} \exp(\text{sim}(S_m^i, S_n^j) / \tau)} \\
\hspace{0pt} + \log \frac{\exp(\text{sim}(S_n^i, S_m^i) / \tau)}{\sum\limits_{j=1}^{T} \exp(\text{sim}(S_n^i, S_m^j) / \tau)} \bigg),
\end{aligned}
\end{equation}

where $\text{sim}(\cdot)$ denotes cosine similarity and $\tau$ is the temperature. This stage promotes coarse-grained cross-lingual alignment without altering the decoder-only architecture.

\paragraph{Stage 2: Paragraph-Level Finetuning.} 
The second stage uses a smaller corpus of paragraph-aligned bilingual texts, consisting of multi-sentence or document-level segments. We apply the same contrastive loss, but compute paragraph embeddings via mean pooling over token-level representations. Despite the smaller scale, this stage is crucial for: (1) extending alignment to longer discourse spans for real-world multilingual applications, and (2) preserving long-text modeling capabilities that may degrade with sentence-only training. This curriculum, moving from short to long text, supports hierarchical multilingual representation learning.

To prevent catastrophic forgetting and retain general language modeling capabilities, we jointly optimize contrastive loss with the NTP loss during both stages:

\begin{equation}
  \mathcal{L}_{\text{PT}}(\theta) = \mathcal{L}_{\text{NTP}}(\theta) + \mathcal{L}_{\text{contrastive}}(\theta).
\end{equation}

\subsection{Cross-lingual Chain-of-Thought (XCoT)}

To further enhance knowledge transfer from high-resource to low-resource languages, we introduce a Cross-lingual Chain-of-Thought (\textsf{XCoT}) prompting strategy during instruction fine-tuning. \textsf{XCoT} guides the model to decompose and solve queries in low-resource languages by leveraging the reasoning capabilities and factual knowledge it has primarily learned in English.

During instruction fine-tuning and inference, \textsf{XCoT} operates in three steps:
\begin{enumerate}
    \setlength{\itemsep}{0pt}
    \setlength{\parskip}{0pt}
    \item \textbf{Reasoning in English.} Given a question in a low-resource language, the model is first prompted to outline the reasoning steps in English, drawing on clearer semantic structures and well-formed logical patterns.
    \item \textbf{Answering in English.} It then generates a concise English answer based on the reasoning trace, capturing the core factual or inferential content.
    \item \textbf{Respond in target language.} Finally, the model returns the answer in the original low-resource language, preserving both accuracy and linguistic appropriateness.
\end{enumerate}

This cross-lingual strategy complements contrastive pretraining by explicitly bridging linguistic gaps during reasoning and generation. Illustrative examples of \textsf{XCoT} are provided in Appendix~\ref{sec:appendix}, Table~\ref{tab:hallucination-examples}.

\section{Experiments}
\label{sec: Experiments}

\subsection{Experimental Settings}
\label{Experimental Settings}
We focus on Singaporean cultural knowledge as our target domain, using English as the high-resource language due to its broad data availability and strong representation in existing LLMs. For multilingual evaluation, we adopt Gemma-7B and LLaMA-3.1-8B, two decoder-only models with multilingual capabilities.

To simulate a realistic scenario where the high-resource language contains domain-specific knowledge, we inject cultural content exclusively into English during instruction fine-tuning. The English instruction fine-tuning dataset is constructed by combining 10,000 samples from CRAFT, a QA dataset focused on Singaporean cultural knowledge \cite{wang2024craftextractingtuningcultural}, with the Alpaca English dataset \cite{alpaca}. From CRAFT, we manually select 500 fact-based QA pairs as the test set and translate them into Chinese and Malay (mid-resource languages) and Tamil (low-resource language), categorized according to their relative representation in the pretraining corpora.

All models are fine-tuned using Low-Rank Adaptation (LoRA) \cite{hu2022lora} for parameter-efficient training. During continued pretraining, we update layers 9–19 for Gemma-7B and layers 11–22 for LLaMA-3.1-8B. In contrast, instruction fine-tuning is applied to all layers in both models. We evaluate hallucination mitigation performance on both models and provide further analysis on the impact of layer-specific tuning. For ablation and exploratory experiments, we use Gemma-7B as the default model. Full dataset construction details are provided in Section~\ref{sec:data_details}.

\subsection{Cross-Lingual Hallucination Analysis}
%\subsection{The extent of hallucination in generative outputs}
We evaluate hallucination rates in non-English QA tasks and compare our proposed method against several baselines. Responses are manually assessed by linguistic experts using authoritative sources. A response is considered \textbf{hallucination-free} if it satisfies:
(1) It accurately and fully answers the question. (2) It contains no factual errors, including in any elaborations.

Representative outputs are shown in Appendix~\ref{sec:appendix}, Table~\ref{tab:hallucination-examples}, covering hallucination-free cases and three common error types: (1) incomplete but seemingly sufficient answers, (2) factually incorrect but topically relevant responses, and (3) irrelevant answers.

Table~\ref{tab:hallucination-free} reports hallucination-free rates for Chinese, Malay, and Tamil. We observe that standard instruction fine-tuning using English domain-specific data fails to generalize effectively to non-English outputs. This underscores a key limitation of current multilingual LLMs: knowledge acquired in high-resource languages does not naturally transfer to low-resource ones. The problem is especially severe for Tamil, where hallucination-free rates are just 1\% (Gemma) and 2\% (LLaMA-3.1). Malay, sharing more lexical overlap with English, achieves the highest baseline performance, 16\% and 18\%, for Gemma and LLaMA-3.1, respectively.

Applying our \textsf{XCoT}strategy yields substantial improvements. Hallucination-free responses increase by 38\% (Gemma) and 36.6\% (LLaMA-3.1) on average across all three languages. The effectiveness of this approach is primarily driven by XCoT’s ability to scaffold the reasoning process using high-resource language knowledge and well-learned reasoning patterns, reducing both factual inaccuracies and incomplete answers.

Integrating \textsf{XCoT} with our cross-lingual contrastive learning objective (\textsf{CL-XCoT}) further amplifies performance gains. The combined framework aligns multilingual representations while enhancing factual transfer from English to other languages. We achieve average increase of hallucination-free rates across both models, reaching 59\% (Chinese), 46\% (Malay), 54\% (Tamil) for Gemma and 65\% (Chinese), 41\% (Malay), 49\% (Tamil) for LLaMA-3.1, significantly outperforming all baselines. Notably, these improvements are consistent across typologically diverse languages, demonstrating the generality of our approach. Paragraph-level contrastive learning (\textsf{CCL-XCoT}) yields additional benefits, with an average improvement of 4\% for Gemma and 4.33\% for LLaMA-3.1 across the three languages. These improvements are primarily attributed to enhanced capabilities in handling long texts and a deeper semantic understanding enabled by longer sequences. These results highlight that with proper alignment and structured reasoning, multilingual LLMs can bridge the semantic and factual gaps between vastly different languages, without needing task-specific data in each target language.

We also benchmark against two competitive baselines:
\begin{itemize}
    \setlength{\itemsep}{0pt}
    \setlength{\parskip}{0pt}
    \item \textbf{CrossIn} \cite{lin2024crossin}: a translation-based fine-tuning approach. Our method surpasses it by over 32\%(Gemma) and 29\%(LLaMA-3.1) in Tamil, indicating that contrastive semantic alignment paired with reasoning transfer is more effective than direct translation in low-resource scenarios.
    \item \textbf{CoV} \cite{dhuliawala2023chain}: a verification-based generation method. Our model outperforms CoV by an average of 53.8\% across all three languages. This suggests that grounding generation in high-resource reasoning may be more effective than post-hoc verification mechanisms in multilingual contexts.
\end{itemize}

To estimate the upper bound of hallucination mitigation, we implement an RAG baseline (Gemma-7B+SFT+RAG), using Sentence-BERT \cite{reimers-2020-multilingual-sentence-bert} to retrieve top-2 relevant paragraphs as in-context prompts. While RAG offers modest improvements in Chinese and Malay, it performs poorly in Tamil. This highlights a critical limitation: retrieval-based methods depend not only on evidence quality, but also on the model’s ability to comprehend and integrate it, which remains fragile in low-resource languages. Our method, in contrast, requires no retrieval infrastructure and delivers stronger, more consistent results.

We further show that larger model scale and more training data fail to reduce hallucinations in low-resource languages, where hallucination-free rates remain low (24\%) despite high rates in English (85\%). Our 8B model outperforms the 70B model in domain-specific QA. See Section \ref{sec:Hallucination-free rate of large LLM} for details.

\begin{table*}[t]
\small
\centering
\begin{minipage}{0.48\linewidth}
\centering
\begin{tabular}{lccc|}
\toprule
\textbf{Method} & \textbf{Chinese} & \textbf{Malay} & \textbf{Tamil} \\
\midrule
Gemma-7B+SFT & 9\% & 16\% & 1\% \\
\midrule
CrossIn\cite{lin2024crossin} & 50\% & 51\% & 28\% \\
CoV\cite{dhuliawala2023chain} & 12\% & 19\% & 3\% \\
Gemma-7B+SFT+RAG & 53\% & 50\% & 30\% \\
\midrule
Gemma-7B+XCoT & 48\% & 55\% & 38\% \\
Gemma-7B+CL-XCoT & 68\% & 62\% & 55\% \\
Gemma-7B+CCL-XCoT & \textbf{71\%} & \textbf{66\%} & \textbf{60\%} \\
\bottomrule
\end{tabular}
\end{minipage}
\hfill
\begin{minipage}{0.48\linewidth}
\centering
\begin{tabular}{lccc}
\toprule
\textbf{Method} & \textbf{Chinese} & \textbf{Malay} & \textbf{Tamil} \\
\midrule
LLaMA-3.1-8B+SFT & 4\% & 18\% & 2\% \\
\midrule
CrossIn\cite{lin2024crossin} & 49\% & 53\% & 26\% \\
CoV\cite{dhuliawala2023chain} & 10\% & 22\% & 4\% \\
LLaMA-3.1-8B+SFT+RAG & 58\% & 52\% & 31\% \\
\midrule
LLaMA-3.1-8B+XCoT & 50\% & 41\% & 43\% \\
LLaMA-3.1-8B+CL-XCoT & 69\% & 59\% & 51\% \\
LLaMA-3.1-8B+CCL-XCoT & \textbf{74\%} & \textbf{63\%} & \textbf{55\%} \\
\bottomrule
\end{tabular}
\end{minipage}
\caption{Hallucination-free rates (\%) in Chinese, Malay, and Tamil under different training strategies. +SFT: direct instruction fine-tuning; +XCoT: apply XCoT during fine-tuning; +CL: integrate sentence-level contrastive learning in pretraining; +CCL: integrate curriculum-based contrastive learning in pretraining. Best results are in \textbf{bold}.} 

\label{tab:hallucination-free}
\end{table*}

\subsection{Impact of Contrastive Learning on Semantic Alignment}
%\subsection{Evaluation of Semantic Space Alignment Degree}

Semantic space alignment is critical for cross-lingual generalization in multilingual LLMs, particularly in zero-shot scenarios \cite{devlin2019bert, conneau2019unsupervised}. To assess this, we adopt the metric from \citet{li2025language}, which measures cosine similarity between average sentence embeddings of semantically aligned sentence pairs across model layers. We use 5,000 such pairs to compute both similarity scores and variance, with lower variance indicating more consistent alignment.

\begin{figure}[t]
  \centering
  \includegraphics[width=0.40\textwidth]{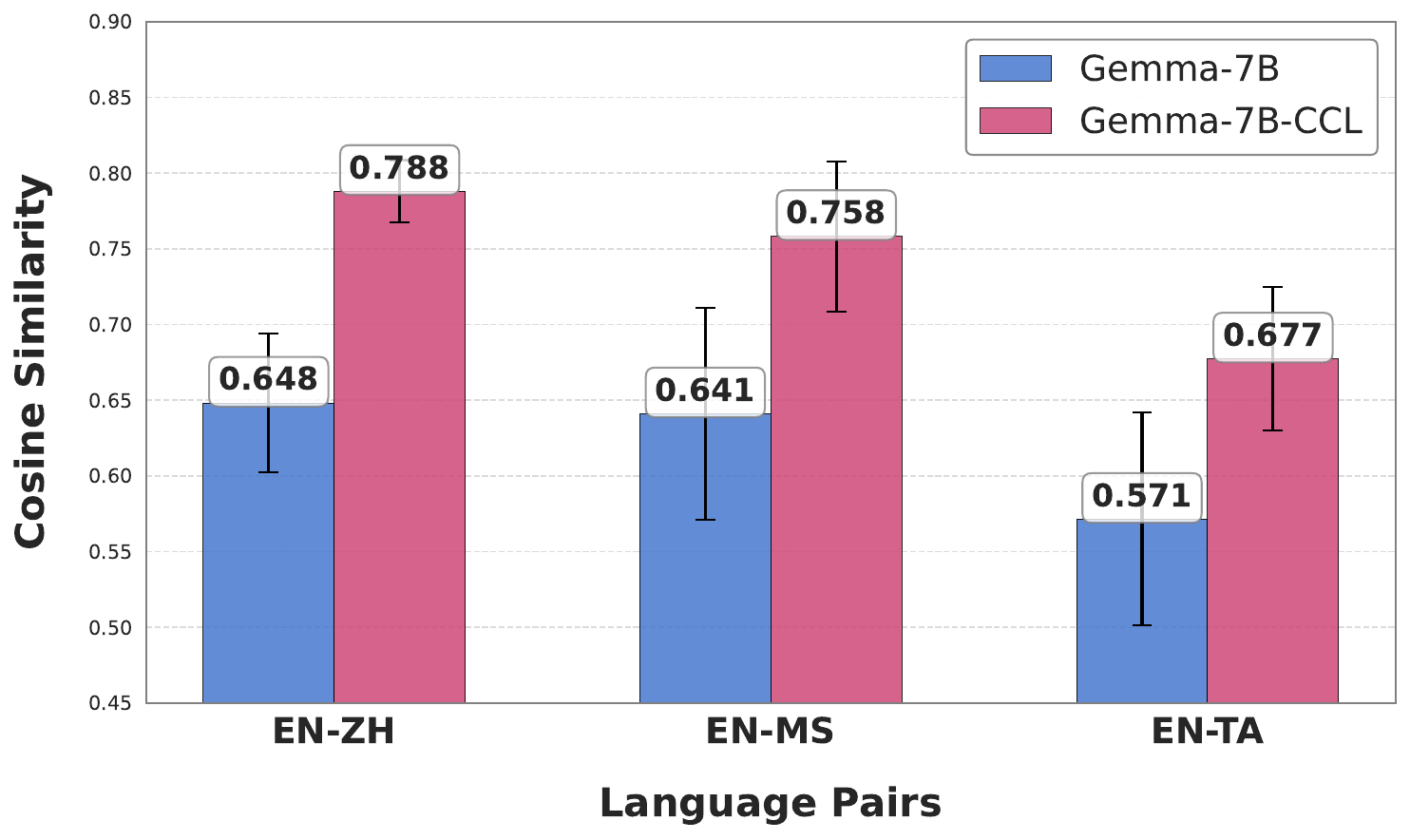}
  \caption{Semantic space alignment across pre-trained models. Bars indicate average cosine similarity between English and target languages; error bars represent sample variance.}
  %\caption{Semantic space alignment of different pre-trained models. The bar chart represents the cosine similarity, and the error bars indicate the variance of the samples used to compute the similarity.}
  \label{fig:similarity}
\end{figure}

As shown in Figure~\ref{fig:similarity}, incorporating contrastive learning during continued pretraining substantially improves semantic alignment. Compared to the baseline Gemma-7B model, our contrastive variant yields similarity gains of 14.0\% (English–Chinese), 11.7\% (English–Malay), and 10.6\% (English–Tamil), along with marked reductions in variance.

These results highlight the role of contrastive learning in improving multilingual representations, especially in underrepresented language pairs. By encouraging the model to anchor semantically equivalent sentences in a shared latent space, our approach enables more reliable transfer of factual and contextual knowledge across languages. This alignment effect underpins downstream improvements in generation quality and hallucination mitigation observed in earlier experiments.

\subsection{Evaluating Cross-Lingual Understanding and Consistency}
%\subsection{Cognitive Consistency Evaluation}

To understand which capabilities are most enhanced by cross-lingual semantic alignment, we categorize NLP tasks into three stages: (1) input understanding, (2) reasoning and inference, and (3) language-specific output. We hypothesize that improved alignment primarily benefits the first two stages, especially for languages with limited pretraining exposure.

We test this by comparing the baseline Gemma-7B model against Gemma-7B+CCL, which incorporates our curriculum-based contrastive learning during continued pretraining. Both models are instruction-tuned on the same dataset. Evaluation is conducted on two multilingual benchmarks: Cross-MMLU (knowledge understanding) and Cross-LogiQA (logical reasoning), drawn from SEA-Eval \cite{wang2023seaeval}, with test sets in English, Chinese, and Malay. We also manually translated and validated Tamil versions of both datasets, resulting in a four-language evaluation.

\begin{figure*}[t]
  \centering
  \includegraphics[width=0.8\textwidth]{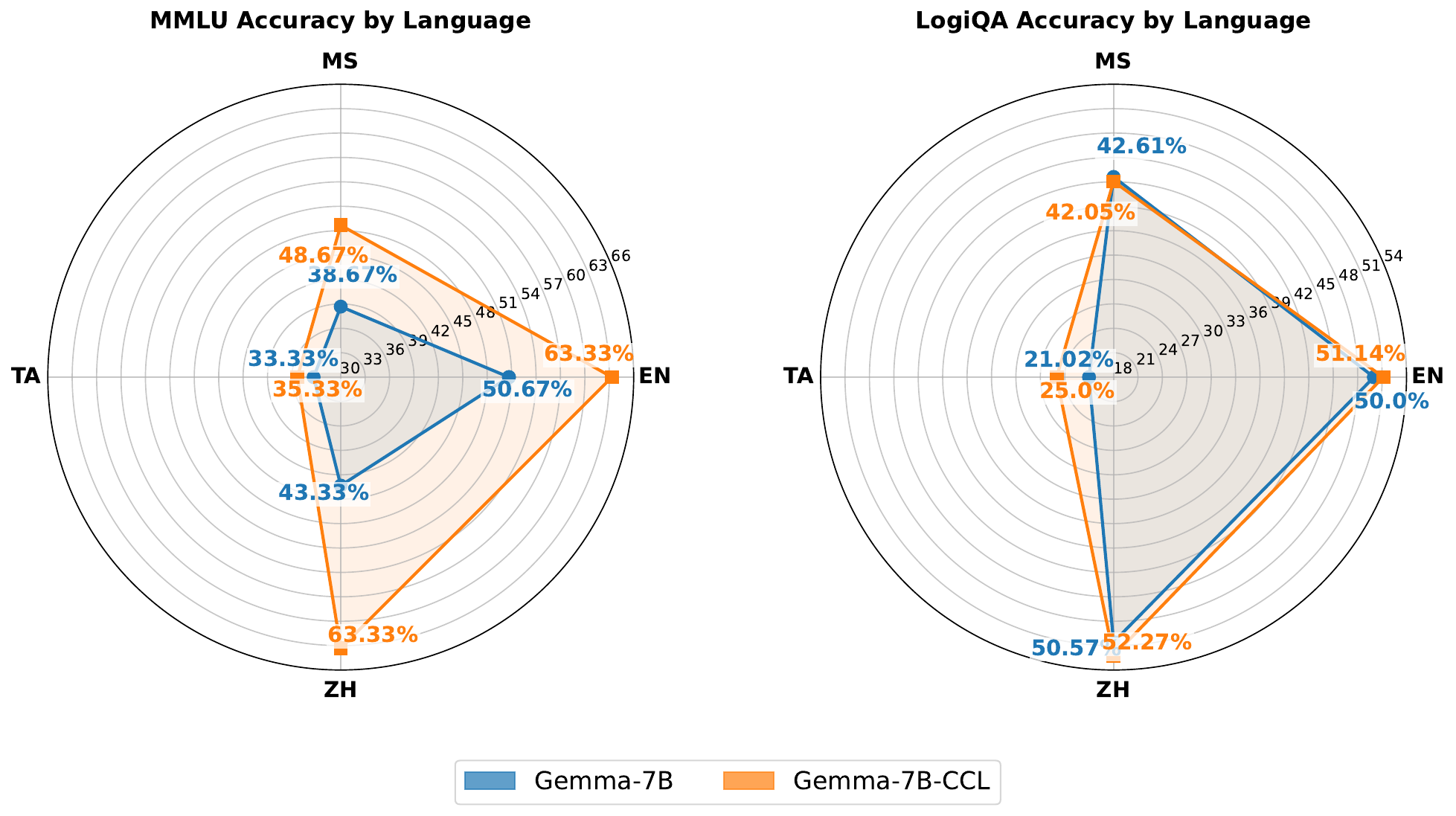}
  \caption{Accuracy comparison on Cross-MMLU and Cross-LogiQA across four languages using Gemma-7B and Gemma-7B+CCL.}
  %\caption{Accuracy Comparison on Cross-MMLU and Cross-LogiQA Across Languages (Gemma-7B vs. Gemma-7B-CCL)}
  \label{fig:answer accuracy}
\end{figure*}

\begin{figure*}[t]
  \centering
  \includegraphics[width=0.8\textwidth]{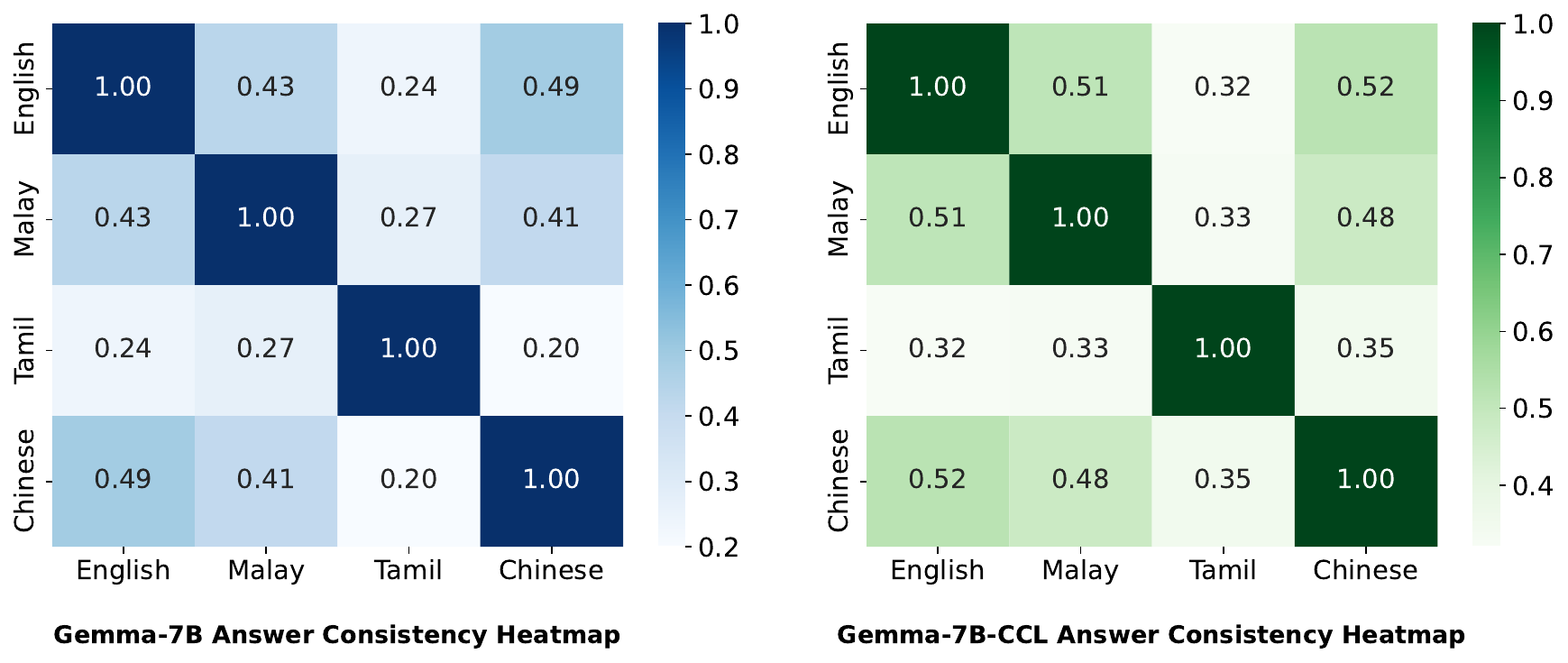}
  \caption{Answer consistency heatmap across language pairs for Gemma-7B and Gemma-7B-CCL.}
  \label{fig:heatmap}
\end{figure*}

As shown in Figure~\ref{fig:answer accuracy}, Gemma-7B+CCL outperforms the baseline across nearly all settings. The largest gains are in Cross-MMLU: Chinese (+20\%) and Malay (+10\%). Cross-LogiQA shows smaller, though generally positive, improvements. These results confirm that semantic alignment significantly boosts multilingual comprehension, while its effect on reasoning is more moderate. Interestingly, English performance also improves (+12.66\% on MMLU), suggesting that multilingual alignment benefits cross-lingual transfer and strengthens English representations, possibly by refining structural understanding and reducing representational redundancy. Tamil, as a low-resource language, shows smaller gains. While MMLU accuracy remains low, Cross-LogiQA improves by 3.98\%. This underscores that alignment helps even when monolingual training is minimal, though limitations in base representation quality cap the benefits.

We also assess cross-lingual consistency, whether the same question yields semantically equivalent answers across languages. As shown in Figure~\ref{fig:heatmap}, consistency with English increases by 8\% for both Tamil and Malay, and 3\% for Chinese. More notably, alignment improves even among non-English pairs (e.g., ZH–TA: +15\%, MS–TA: +6\%, ZH–MS: +7\%) despite no direct training signal between them. These findings reveal that English can act as a pivot language, indirectly aligning low-resource language pairs through shared semantic structure. This opens up promising directions for multilingual model training without relying on costly parallel corpora.

\subsection{Layer-Wise Analysis of Cross-Lingual Knowledge Transfer}
%\subsection{Layer-wise Analysis of knowledge transfer}

\begin{figure}[t]
  \centering
  \includegraphics[width=0.40\textwidth]{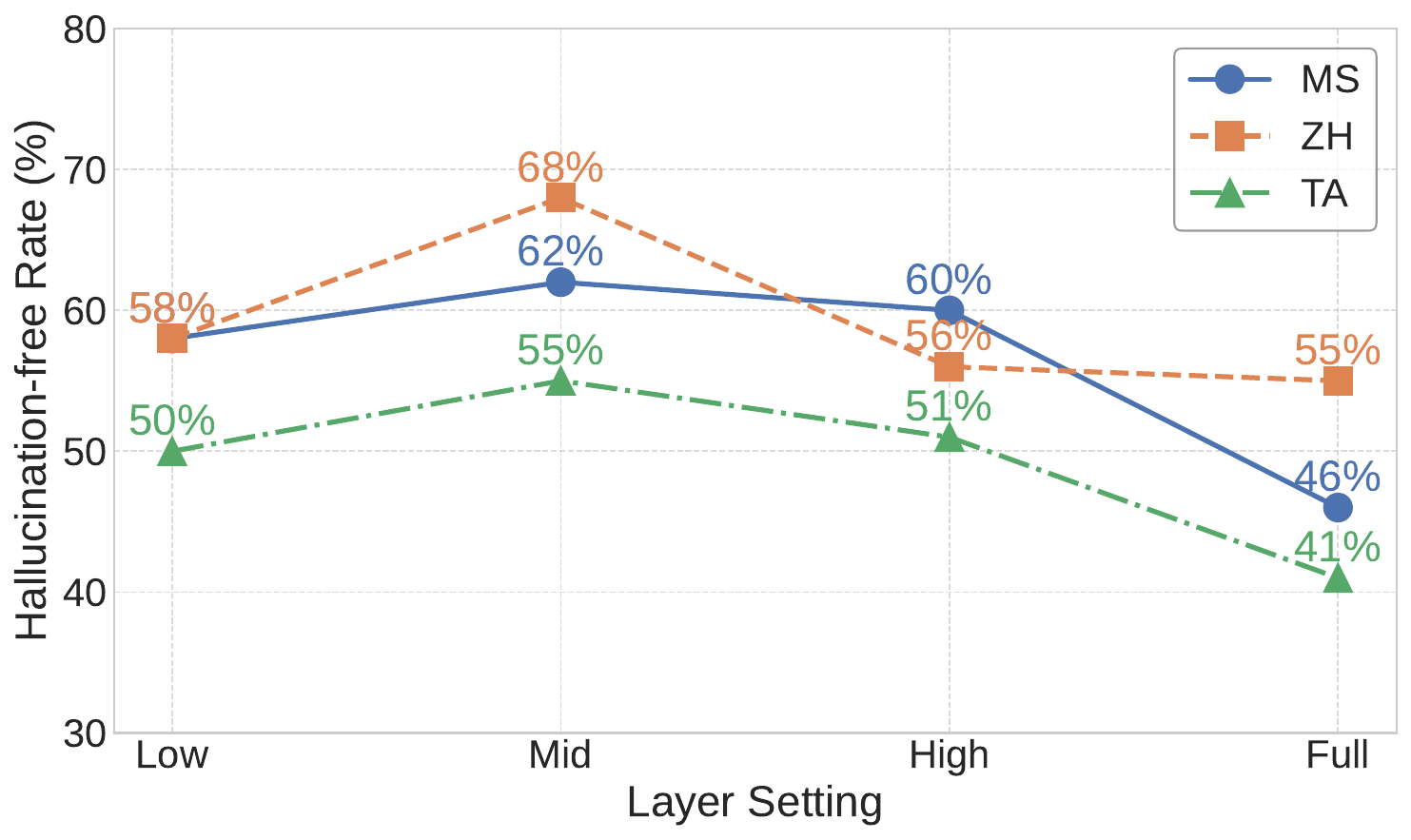}
  \caption{Hallucination-free rates by language across different layer-wise fine-tuning settings.}
  %\caption{Hallucination-free Rate by Language and Layer Setting}
  \label{fig:hall_layers}
\end{figure}

To examine which layers are most responsible for cross-lingual knowledge transfer, we divide the Gemma-7B architecture into three segments: low-level (layers 0–8), mid-level (9–19), and high-level (20–27). We selectively fine-tune each segment and measure its effect on hallucination-free rates across Chinese, Malay, and Tamil.

As shown in Figure~\ref{fig:hall_layers}, fine-tuning only the \textbf{mid-level layers} achieves the best performance across all languages. High-level tuning ranks second, while full-model fine-tuning surprisingly underperforms both. These findings suggest that mid-level layers play a pivotal role in aligning semantic representations across languages, likely functioning as the core bridge between surface-level linguistic features and higher-order reasoning.

In contrast, fine-tuning low-level layers leads to a noticeable performance drop. We hypothesize that these layers encode language-specific syntactic and morphological patterns; altering them may distort the model’s capacity to process diverse linguistic inputs. Worse, updates in these early layers can propagate undesirable shifts to downstream layers, disrupting the semantic abstraction necessary for effective knowledge transfer. These results highlight the importance of targeted mid-layer adaptation in multilingual transfer settings—offering a more efficient and stable alternative to full-model tuning.

\section{Case Study}
\label{sec:case_studies}

To better understand hallucination behavior across languages, we conduct a targeted case study comparing model outputs in medium-resource languages (Chinese and Malay) and a low-resource language (Tamil). After direct instruction fine-tuning, we observe distinct error profiles. In Chinese and Malay, hallucinations typically fall into two categories: unnecessary elaborations beyond the scope of the prompt, and misinterpretations of culturally specific or domain-relevant terms, such as \textit{Cendol} or the \textit{Home Ownership Scheme}. In contrast, hallucinations in Tamil are predominantly severe, with over 90\% of cases producing responses that are entirely unrelated to the input question, highlighting the fragility of the model’s representations in truly low-resource settings.

Introducing the \textsf{XCoT} strategy leads to substantial improvements across all languages. For Chinese and Malay, \textsf{XCoT} reduces over-generation errors by encouraging more focused and structured reasoning. However, challenges with domain-specific terminology persist, as these terms are often underrepresented in the training data. In Tamil, \textsf{\textsf{XCoT}} improves response relevance by providing a clearer reasoning scaffold, but some hallucinations remain, often manifesting as topic drift or vague associations. When \textsf{XCoT} is combined with cross-lingual contrastive learning, we observe further improvements in all three languages, particularly in handling terminology. In medium-resource settings, hallucinations are reduced to minor factual inaccuracies, such as incorrect years or road names, indicating a transition from gross misunderstanding to finer-grained errors. In Tamil, hallucinations become more diverse but notably less frequent, suggesting better control over output semantics despite limited language-specific training signals.

These findings highlight the importance of tailoring multilingual LLM interventions based on language resource availability. For medium-resource languages, enhancing factual precision and grounding in culturally specific terminology should be prioritized. For low-resource languages like Tamil, future improvements will likely require more robust semantic alignment mechanisms and adaptive reasoning strategies to address a broader range of hallucination types and maintain answer relevance. 
% Additional qualitative examples are included in Section \ref{sec:case_study} to further illustrate these patterns.

\section{Conclusion}

This paper present \textsf{CCL-XCoT}, an efficient two-stage framework for cross-lingual knowledge transfer in multilingual LLMs. Our approach combines curriculum-based contrastive learning during continued pretraining with \textsf{XCoT} strategy during instruction fine-tuning. Together, these components align semantic representations across languages and scaffold reasoning through high-resource language traces, substantially reducing hallucinations in low-resource question answering tasks.

Extensive experiments show that contrastive learning improves semantic alignment and consistency, while \textsf{XCoT} enables more accurate and faithful generation across diverse language pairs. Our layer-wise analysis reveals that cross-lingual transfer is most effective when concentrated in mid-level layers, offering a practical path toward efficient multilingual adaptation. Case studies further demonstrate how hallucination patterns vary by resource level, highlighting the need for differentiated strategies.

Looking ahead, future work may explore integrating multilingual retrieval augmentation, expanding \textsf{XCoT}-style reasoning to additional modalities, and developing lightweight adapters to support broader cross-lingual generalization, especially in zero-shot or resource-sparse scenarios.

\section*{Limitations}

Although our proposed training strategy demonstrates high effectiveness in promoting cross-lingual sharing of factual knowledge, its impact on enhancing capabilities for complex logical reasoning tasks is limited. Furthermore, while the strategy primarily focuses on reducing the overall hallucination rate, it lacks effective suppression mechanisms for more fine-grained hallucination types, such as those involving years, designer names, or road numbers, which remain prevalent even in high-resource languages.

% \bibliographystyle{plainnat}
% \bibliography{anthology,custom}
\bibliography{custom}

\clearpage

\appendix

\section{Appendix}
\label{sec:appendix}

\subsection{Gold Answer and Hallucination Examples}
The hallucination-free cases and three common error types are shown in Table \ref{tab:hallucination-examples}.

\begin{table*}[htbp]
\centering
\small
\renewcommand{\arraystretch}{1.4}

% 表头
\begin{tabularx}{\textwidth}{X|X|X|X}
\hline
\textbf{Golden Answer} & \textbf{Incomplete Response} & \textbf{Factual Error} & \textbf{Irrelevant response} \\
\hline\hline
% 内容
Kampong Spirit refers to the strong sense of community, mutual support, and togetherness found in traditional village life. It originated in early Singapore, where residents of close-knit kampongs relied on one another for daily needs and survival. &
Kampong Spirit refers to the strong sense of community, mutual support, and togetherness found in traditional village life. &
Kampong Spirit refers to the strong drive to compete and persevere, rooted in the experiences of early Chinese immigrants to Singapore who strived to improve their lives through hard work and mutual competition. &
The Kiasu spirit in Singapore reflects a fear of missing out and a strong desire to stay ahead of others. It drives people to be highly competitive, hardworking, and resourceful in all aspects of life. \\
\hline\hline
% 正确性行
\textcolor{green}{\checkmark} & \textcolor{red}{\ding{55}} & \textcolor{red}{\ding{55}} & \textcolor{red}{\ding{55}} \\
\hline
\end{tabularx}

\caption{Example responses to question \textit{What is Singapore's Kampong Spirit? And what are its origins?} for hallucination evaluation.}
\label{tab:hallucination-examples}
\end{table*}

\subsection{Comparison of Hallucination-Free Rate with Large-Scale LLMs}
\label{sec:Hallucination-free rate of large LLM}

We demonstrate that scaling model size and expanding training corpora do not effectively reduce hallucinations in low-resource languages. While English achieves a high hallucination-free rate of 85\%, low-resource languages persist at a low 24\%. As presented in Table \ref{table: Comparison of hallucination-free rate with 70B}, our approach, applied to LLaMA-3.1-8B, consistently surpasses LLaMA-3.1-70B-Instruct in domain-specific question-answering tasks related to Singaporean culture, yielding improvements of 32\%, 12\%, and 31\% in Chinese, Malay, and Tamil, respectively. These findings indicate that merely increasing model size and training data is insufficient for enhancing cross-lingual knowledge generalization, particularly for low-resource languages. In the future, we will apply our method to larger-scale LLMs once sufficient computational resources are available.

\begin{table*}[ht]
\centering
\begin{tabular}{lcccc}
\toprule
\textbf{Model} & \textbf{English} & \textbf{Chinese} & \textbf{Malay} & \textbf{Tamil} \\
\midrule
Llama-3.1-70B-Instruct  & 85\% & 42\% & 50\% & 24\% \\
LLaMA-3.1-8B+CCL-XCoT & \textbf{94\%} & \textbf{74\%(+32\%)} & \textbf{63\%(+13\%)} & \textbf{55\%(+31\%)} \\
\bottomrule
\end{tabular}
\caption{Comparison of hallucination-free rate between our method and large-scale LLMs. Best results are in \textbf{bold}.}
\label{table: Comparison of hallucination-free rate with 70B}
\end{table*}

\subsection{Data Details}
\label{sec:data_details}

In the contrastive learning experiments, we selected a subset of sentence pairs from the WMT-2024 dataset as training data for the EN-ZH language pair. To increase data diversity and ensure sufficient semantic distinction among negative samples within each batch, we extracted 100K sentence pairs with varying lengths and degrees of semantic similarity. For the EN-MS and EN-TA pairs, due to the absence of high-quality open-source datasets, we relied on data collected internally by our team. The data selection criteria and overall volume were kept consistent with those used for EN-ZH. 

For the Chain-of-Thought (CoT) data used during the instruction fine-tuning phase, we extracted 3,000 QA pairs from the CRAFT dataset that are non-overlapping with the test set, both in terms of surface question formulations and the underlying knowledge content they assess. CoT annotations were generated using ChatGPT-4o based on these QA pairs through a 3-shot in-context learning approach. The generated data was manually verified by linguistic experts to ensure quality.

The semantic space alignment test data consists of two parts: the complete FLORES-200 dataset\cite{nllb2022} and a subset sampled from our self-collected data. Each test set for a language pair contains 5,000 sentence pairs, with no overlap between the test and training data.

\begin{table*}[t]
\centering
\small
\renewcommand{\arraystretch}{1.2}
\begin{tabular}{p{3cm}p{2.5cm}p{2cm}p{1.8cm}p{2.5cm}p{1.8cm}}
\hline
\textbf{Task Name} & \textbf{\makecell[l]{Language /\\Language Pairs}} & \textbf{Data Volume} & \textbf{Data Description} & \textbf{Data Source} & \textbf{Human-verified} \\
\hline
\multirow{3}{2.8cm}{Sentence-level 
contrastive learning} & EN-ZH & 100K & Sentence Pair & WMT2024 & No\\
 & EN-MS & 100K & Sentence Pair  & self-collected data & No\\ 
 & EN-TA & 100K & Sentence Pair  & self-collected data & No\\
\hline
\multirow{3}{2.8cm}{Paragraph-level 
contrastive learning} & EN-ZH & 10K & Paragraph Pair & WMT2024 & No \\
 & EN-MS & 10K & Paragraph Pair  & self-collected data & No\\ 
 & EN-TA & 10K & Paragraph Pair  & self-collected data & No\\
\hline
\multirow{3}{2.8cm}{CoT Instruction Finetune} & EN-ZH & 3K & Question-Answering & GPT-generated(3-shots) & Yes  \\
 & EN-MS & 3K & Question-Answering  & GPT-generated(3-shots) & Yes\\ 
 & EN-TA & 3K & Question-Answering  & GPT-generated(3-shots) & Yes\\
 \hline
\multirow{3}{2.8cm}{Semantic space alignment test data} & EN-ZH & 5K & Sentence Pair  & FLORES-200+self-collected & No  \\
 & EN-MS & 5K & Sentence Pair  & FLORES-200+self-collected & No\\ 
 & EN-TA & 5K & Sentence Pair  & FLORES-200+self-collected & No\\

\hline
\end{tabular}
\caption{Task Details and Data Statistics}
\label{tab:task_details_no_subtask}
\end{table*}

\end{CJK}

\end{document}